\documentclass[english, 13pt]{article}
\usepackage{textcomp}
\usepackage{mathrsfs}
\usepackage{mathtools}
\usepackage{amsmath}
\usepackage{amsthm}
\usepackage{amssymb}
\usepackage{dsfont}
\usepackage{esint}
\usepackage{listings}
\usepackage{float}
\usepackage{caption} 
\usepackage{subcaption} 
\usepackage{xspace}
\usepackage{authblk}
\usepackage[dvipsnames]{xcolor}
\usepackage[colorlinks=true,citecolor=blue,linkcolor=blue,urlcolor=blue]{hyperref}
\usepackage{cleveref}
\usepackage[ruled, vlined, linesnumbered]{algorithm2e}
\usepackage[
margin=1.5cm,
includefoot,
footskip=30pt,
]{geometry}
\usepackage[
backend=biber,
style=authoryear,
sorting=nyt,
maxcitenames=2,
maxbibnames=99,
url=false]{biblatex}
\usepackage{scrextend}
\usepackage{color}
\deffootnote[1em]{1em}{1em}{%
   \textsuperscript{\thefootnotemark}%
}
\crefname{algocf}{algorithm}{algorithms}
\newcommand{\pz}{\mathbf{p}(0)}
\newcommand{\pt}{\mathbf{p}(t)}

\newcommand{\dth}[1]{\frac{\partial{#1}}{\partial{\theta}}}
\newcommand{\dths}[1]{#1'}

\newcommand{\expm}[1]{\exp\!\left(#1\right)\!}
\DeclareMathOperator{\Id}{I}

\title{Differentiated uniformization: A new method for inferring Markov chains on combinatorial state spaces including stochastic epidemic models}

\date{\vspace{-5ex}}

\author[1]{Kevin Rupp\footnote{These authors contributed equally.}} 
\author[1]{Rudolf Schill\protect\footnotemark[1]\footnote{Correspondence: Rudolf.Schill@klinik.uni-regensburg.de, Rainer.Spang@klinik.uni-regensburg.de}} 
\author[1]{Jonas Süskind\protect\footnotemark[1]} 
\author[2]{Peter Georg}
\author[3]{Maren Klever}
\author[1]{Andreas Lösch}
\author[3]{\authorcr Lars Grasedyck}
\author[2]{Tilo Wettig}
\author[1]{Rainer Spang\protect\footnotemark[2]}
\affil[1]{Department of Statistical Bioinformatics, University of Regensburg,  93040 Regensburg, Germany}
\affil[2]{Department of Physics, University of Regensburg, 93040 Regensburg, Germany}
\affil[3]{Institut für Geometrie und Praktische Mathematik, RWTH Aachen University, 52062 Aachen, Germany}

\begin{document}

\maketitle

\begin{abstract}
\noindent\textbf{Motivation:} We consider continuous-time Markov chains that describe the stochastic evolution of a dynamical system by a transition-rate matrix $Q$ which depends on a parameter $\theta$. 
Computing the probability distribution over states at time $t$ requires the matrix exponential $\exp(tQ)$, and inferring $\theta$ from data requires its derivative $\partial\exp\!(tQ)/\partial\theta$. 
Both are challenging to compute when the state space and hence the size of $Q$ is huge. This can happen when the state space consists of all combinations of the values of several interacting discrete variables. Often it is even impossible to store $Q$.
However, when $Q$ can be written as a sum of tensor products, computing $\exp(tQ)$ becomes feasible by the uniformization method, which does not require explicit storage of $Q$. 

\noindent\textbf{Results:} Here we provide an analogous algorithm for computing $\partial\exp\!(tQ)/\partial\theta$, the \emph{differentiated uniformization method}.
We demonstrate our algorithm for the stochastic SIR model of epidemic spread, for which we show that $Q$ can be written as a sum of tensor products. We estimate monthly infection and recovery rates during the first wave of the COVID-19 pandemic in Austria and quantify their uncertainty in a full Bayesian analysis.

\noindent\textbf{Availability:} Implementation and data are available at \url{https://github.com/spang-lab/TenSIR}.
\end{abstract}

\section{Introduction}
Predicting the time evolution of complex dynamical systems has a wide range of applications in medicine and public health.
One of them is the SIR model of epidemic spread, which describes a population by the numbers of susceptible ($S$), infected ($I$) and recovered ($R$) people.
Until recently the SIR model has been approximated deterministically \parencite{kendrick27} and was considered computationally intractable in its stochastic formulation \parencite{kendrick25}. 
The stochastic SIR model is a continuous-time Markov chain (CTMC) in which infections happen randomly with a rate proportional to $S$ and proportional to $I$ \parencite{allen17}. 
The state of the system at a given time is thus fully specified by the combination of $S$ and $I$. Since infections happen randomly one must keep track of a huge number of probabilities, one for every possible state.
For example, the Austrian population of 9 million people can go through roughly $9\text{ million}\times9\text{ million} = 81\text{ trillion}$ possible states during the course of an epidemic.

More generally, we consider CTMCs that describe the evolution of a probability distribution $\pt$  over a huge discrete state space according to the Kolmogorov forward equation
\begin{align}
\frac{\mathrm{d}\pt}{\mathrm{d}t} = Q \pt \quad\text{ with solution }\quad \pt = \expm{tQ} \pz.
\end{align}
Here $Q$ is the transition-rate matrix and $\expm{tQ}$ is the matrix exponential. For an SIR model of the Austrian population $Q$ has $81\text{ trillion}\times81\text{ trillion}$ entries, and naively computing the matrix exponential requires on the order of $81\text{ trillion}\times81\text{ trillion}\times81\text{ trillion}$ operations \parencite{moler03}, which is practically impossible.

Even more dauntingly, when $Q$ depends on an unknown parameter $\theta$, such as the infection or recovery rate in the SIR model, we must first infer $\theta$ from data by maximizing its likelihood or by sampling from its posterior in a full Bayesian analysis.
This typically requires the derivative of the matrix exponential $\partial \expm{tQ}\pz/\partial\theta$ in order to compute $\partial \pt/\partial\theta$.
However, \textcite{ho18} have recently provided an algorithm that solves the Kolmogorov equation in the Laplace domain and evaluates the inverse Laplace transform numerically, thus avoiding the matrix exponential. Their algorithm is applicable to systems where each discrete variable increases monotonically. This includes the SIR model,\footnote{By changing variables from susceptibles and infected to infections and recoveries.} for which their algorithm scales quadratically in the population size. 

Here, we provide an alternative algorithm that  directly computes $\expm{tQ}$ and $\partial
\expm{tQ}/\partial\theta$. For the SIR model it scales cubically in the population size but is still practical. Importantly, our approach is applicable to a broader class of CTMCs with large state spaces that arise from interacting discrete variables, without requiring monotonicity.
For example, in tumor progression models the states are combinations of possible mutations (\cite{beerenwinkel09}, \cite{schill19}), in stochastic neural networks the states are activation patterns of neurons \parencite{yamanaka97}, in predator-prey dynamics they are joint population sizes of interacting species \parencite{owen14}, or in chemical reaction networks they are joint counts of chemical species \parencite{wolf19}.

For many of these models $Q$ can be written as a sum of tensor products \parencite{buchholz99}. We provide such a representation for the stochastic SIR model. To the best of our knowledge, this representation is novel.
We use it for matrix-vector products that do not require explicit storage of $Q$ \parencite{buis96} and make computation of the matrix exponential tractable via the uniformization method \parencite{grassmann77}. A similar approach by \textcite{sherlock21} exploits the sparsity of $Q$. We extend the uniformization method and provide an analogous algorithm that also computes the derivative of the matrix exponential.
Finally, we use Hamiltonian Monte Carlo sampling to provide a full Bayesian analysis of the first wave of the COVID-19 pandemic for the Austrian population, shedding new light on the uncertainties associated with the estimation of infection and recovery rates.

\section{Differentiated uniformization for parameter estimation}
A discrete-state, continuous-time Markov chain (CTMC) describes probability distributions $\pt \in \mathbb{R}^{|X|}$ over a state space $X$, where an entry $\pt_x$ denotes the probability that the CTMC is in state $x \in X$ at time $t\in [0, \infty)$. Its change over time is governed by the Kolmogorov forward equation
\begin{align} \label{Kolmogorov}
\frac{\mathrm{d} \pt}{\mathrm{d} t} = Q \pt
\end{align}
with transition-rate matrix $Q \in \mathbb{R}^{|X| \times |X|}$, where an off-diagonal entry $Q_{y,x}$ is the transition rate from state $x \in X$ to state $y \in X$ and diagonal entries are set such that columns sum to zero.

The solution to \cref{Kolmogorov} is given by the matrix exponential 
\begin{align}\label{eq:ptseries}
\pt = \expm{tQ} \pz &= \sum_{n=0}^{\infty} \frac{t^n}{n!} Q^n \pz,
\end{align}
which could be approximated in principle by terminating after a finite number of terms. However, catastrophic cancellations occur \parencite{moler03} due to the fact that $Q$ has negative entries and negative eigenvalues.\footnote{Negative entries directly lead to cancellations in \cref{eq:ptseries}. If  \cref{eq:ptseries} is transformed to the eigenbasis of $Q$, negative eigenvalues of $Q$ lead to cancellations as well.} The uniformization method \parencite{grassmann77} addresses this problem by introducing a strictly nonnegative matrix 
\begin{align}
P := \frac{1}{\gamma}Q+\Id \text{\quad for some } \gamma \geq \underset{x}{\max}|Q_{x,x}|
\end{align}
such that
\begin{align} 
\pt = \expm{tQ} \pz 
&= \expm{\gamma t (-\Id+ P)} \pz \nonumber\\
&= \expm{- \gamma t\Id} \expm{\gamma tP}  \pz \nonumber\\
&= \sum_{n=0}^{\infty} e^{-\gamma t} \frac{(\gamma t)^n}{n!}  P^n \pz \label{UnifSeries}
\end{align}
does not suffer from cancellations. $P$ can be viewed as the transition probability matrix of a discrete-time Markov chain where the number of transitions is a Poisson-distributed random variable with mean $\gamma t$.

Using the recursions
\begin{align} 
P^n &= PP^{n-1}, \label{recPow}\\
\frac{(\gamma t)^n}{n!} &= \frac{\gamma t}{n} \frac{(\gamma t)^{n-1}}{(n-1)!}, \label{recFrac}
\end{align}
$\pt$ can be computed according to \cref{UnifSeries} by \cref{algo1} \parencite{grassmann77}. 
Note that $P^n \pz$ sums to 1 and hence \cref{UnifSeries} sums to less than 1 when terminated after a finite number of terms. The algorithm stops once this probability mass defect 
\begin{align} \label{eq:MassDefect}
1 - \sum_{n=0}^{m} e^{-\gamma t}\frac{(\gamma t)^n}{n!}
\end{align}
is smaller than a preset tolerance $\varepsilon$. The required number $m$ of iterations is in $\mathcal{O}(\gamma)$ \parencite{reibman88} and can be determined, e.g., using the numerically robust method by \textcite{sherlock21}.

In this paper we are interested in statistical models where $Q$ depends on a parameter $\theta$ that we want to estimate from data by maximizing its likelihood or by sampling from its posterior.
To this end, we propose a novel algorithm for computing the derivative
\begin{align} 
\dth{\pt}  &= \dth{\expm{tQ}\pz} \nonumber\\
&= \dth{} \left( \sum_{n=0}^{\infty}  e^{-\gamma t}\frac{(\gamma t)^n}{n!} P^n \pz \right)\nonumber\\ 
&= \sum_{n=0}^{\infty} e^{-\gamma t}\frac{(t\gamma)^n}{n!}  \dth{P^n} \pz + e^{-\gamma t} \dth{\gamma} \left(-\frac{t^{n+1}\gamma^n}{n!} + \frac{t^n\gamma^{n-1}}{(n-1)!}\right) P^n \pz \nonumber\\
&=\sum_{n=0}^{\infty} e^{-\gamma t}\frac{(t\gamma)^n}{n!}\left(\frac{\partial P^n}{\partial \theta} \pz + \dth{\gamma}\left(\frac{n}{\gamma}-t\right)P^n \pz \right) \label{DiffUnifSeries}
\end{align}
building on the uniformization method. We use the recursions (\ref{recPow}), (\ref{recFrac}) and additionally
\begin{align}
\dth{P^{n}} &= \dth{P}P^{n-1}+ P \dth{P} P^{n-2}  + \ldots +  P^{n-2} \dth{P}P+ P^{n-1} \dth{P} \notag\\ \label{rec1}
&=\dth{P}P^{n-1} + P \left( \dth{P} P^{n-2}  + \ldots +  P^{n-3} \dth{P}P+ P^{n-2} \dth{P} \right) \notag\\
&=\dth{P}P^{n-1} + P \left(  \dth{P^{n-1}} \right)
\end{align}
to compute $\pt':=\partial \pt/\partial\theta$ according to \cref{DiffUnifSeries} by \cref{algo2}. 

\begin{centering}
	\begin{minipage}[t]{.38\textwidth}
		\vspace{0pt}  
		\begin{algorithm}[H]
			\caption{Uniformization}
		    \label{algo1}
			\SetKwInOut{Input}{input}\SetKwInOut{Output}{output}
			\Input{$\pz, t, P, \gamma, \varepsilon$} 
			\Output{$\pt$}
			$n\, \gets 0$\\
			$w \gets 1$\\
			$\pt \gets \mathbf{0}$\\
			$\mathbf{q} \gets \pz$\\
			\Repeat{$1-\vert\pt\vert_1 < \varepsilon$}{
				$\pt \gets \pt + e^{-\gamma t}w \mathbf{q}$\\
				$n \gets n+1$\\
				$\mathbf{q} \gets P \mathbf{q}$\\
				$w \gets \frac{\gamma t}{n}w$	
			}
			\Return $\pt$ \\
		\end{algorithm}
	\end{minipage}\qquad
	\begin{minipage}[t]{.51\textwidth}
		\vspace{0pt}
		\begin{algorithm}[H]
			\caption{Differentiated Uniformization}
			\label{algo2}
			\SetKwInOut{Input}{input}\SetKwInOut{Output}{output}
			\Input{$\pz, t, P, \dths{P}, \gamma, \dths{\gamma}, \varepsilon$} 
			\Output{$\pt,\dths{\pt}$}
			$n\, \gets 0$\\
			$w \gets 1$\\
			$\pt\phantom{'} \gets \mathbf{0}$\\
			$\dths{\pt} \gets \mathbf{0}$\\
			$\mathbf{q}\phantom{'} \gets \pz$\\
			$\mathbf{q}' \gets \mathbf{0}$\\
			\Repeat{$1-\vert\pt\vert_1 < \varepsilon$}{
				$\pt\phantom{'} \gets \pt\phantom{'} + e^{-\gamma t}w \mathbf{q}$\\
				$\dths{\pt} \gets \dths{\pt} + e^{-\gamma t}w \left( \mathbf{q}' +  \dths{\gamma}\left(\frac{n}{\gamma}-t\right) \mathbf{q} \right)$ \\
				$n\phantom{'} \gets n+1$\\
				$\mathbf{q}' \gets \dths{P} \mathbf{q} + P \mathbf{q}'$\\
				$\mathbf{q}\phantom{'} \gets P \mathbf{q}$\\
				$w\,\gets \frac{\gamma t}{n}w$	
			}
			\Return $\pt,\dths{\pt}$
		\end{algorithm}
	\end{minipage}
\end{centering}

Applying differentiated uniformization for a particular statistical model requires the scalars 
\begin{align}
\gamma \geq \underset{x}{\max}|Q_{x,x}| \quad\text{ and }\quad \gamma' := \dth{\gamma},
\end{align}
where a generic choice for $\gamma$ can be the $2$-norm of the diagonal of $Q$ or any  $p$-norm with even $p$. It also requires the operators
\begin{align}
P = \frac{1}{\gamma}Q+\Id  \quad\text{ and }\quad P' := \dth{P} &= -\frac1{\gamma^2}\dth{\gamma}Q + \frac1\gamma\dth{Q}.
\end{align}
Crucially, these operators are only needed for matrix-vector products in lines 11 and 12  of \cref{algo2} and do not need to be stored explicitly. This makes our method especially useful for models where $Q$ is large but has a compact representation as a sum of tensor products, which allows one to cheaply compute matrix-vector products \parencite{buis96}.

\paragraph{}
Differentiated uniformization thus opens the door to \textbf{parameter inference} for CTMCs on huge discrete state spaces. Let $\{x_1, \ldots, x_K\}$ be observations of the Markov chain at corresponding time points $\{t_1, \ldots, t_K\}$. We represent each data point by an empirical probability distribution $\boldsymbol{\delta}(t_k) \in \mathbb{R}^{\vert X\vert}$, where $\boldsymbol{\delta} (t_k)_{x_k}=1$ and all other entries are zero.
The likelihood of $\theta$ for a single observation of state $x_k$ at time $t_k$ with $k>1$ is
\begin{align}
\mathbf{p}(t_k)_{x_k} \text{, where }\mathbf{p}(t_k) =
\exp\!\big((t_{k}-t_{k-1}) Q\big)
\boldsymbol{\delta}(t_{k-1}).
\end{align}
The log-likelihood for the whole data set,
\begin{equation}
\ell(\theta) =\sum_{k=2}^{K}\log(\mathbf{p}(t_k)_{x_k}), 
\end{equation}
can be maximized using its derivative
\begin{equation}\label{eq:LL_}
\dth{\ell(\theta)} =\sum_{k=2}^{K}\frac{\mathbf{p}(t_k)_{x_k}'}{\mathbf{p}(t_k)_{x_k}},
\end{equation}
for example by gradient ascent. This derivative can also be used for sampling a posterior distribution of $\theta$ in a full Bayesian model using a Hamiltonian Monte Carlo method \parencite{gelman04}. 

\section{Modeling epidemic spread}
The most basic models of epidemic spread are SIR models, which describe the numbers  of susceptible ($S$),  infected/infectious ($I$) and recovered ($R$) people during an epidemic in a closed population of constant size $N$. 
There is a widely known deterministic and a lesser-known stochastic variant of the SIR model \parencite{allen17}. The latter involves a huge discrete state space and is therefore challenging to use. However, it allows for uncertainty quantification in its dynamics and in inferred parameters.
\begin{figure}[H]
	\centering
	\begin{subfigure}[t]{.44\textwidth}
		\centering
		\includegraphics[width=\linewidth]{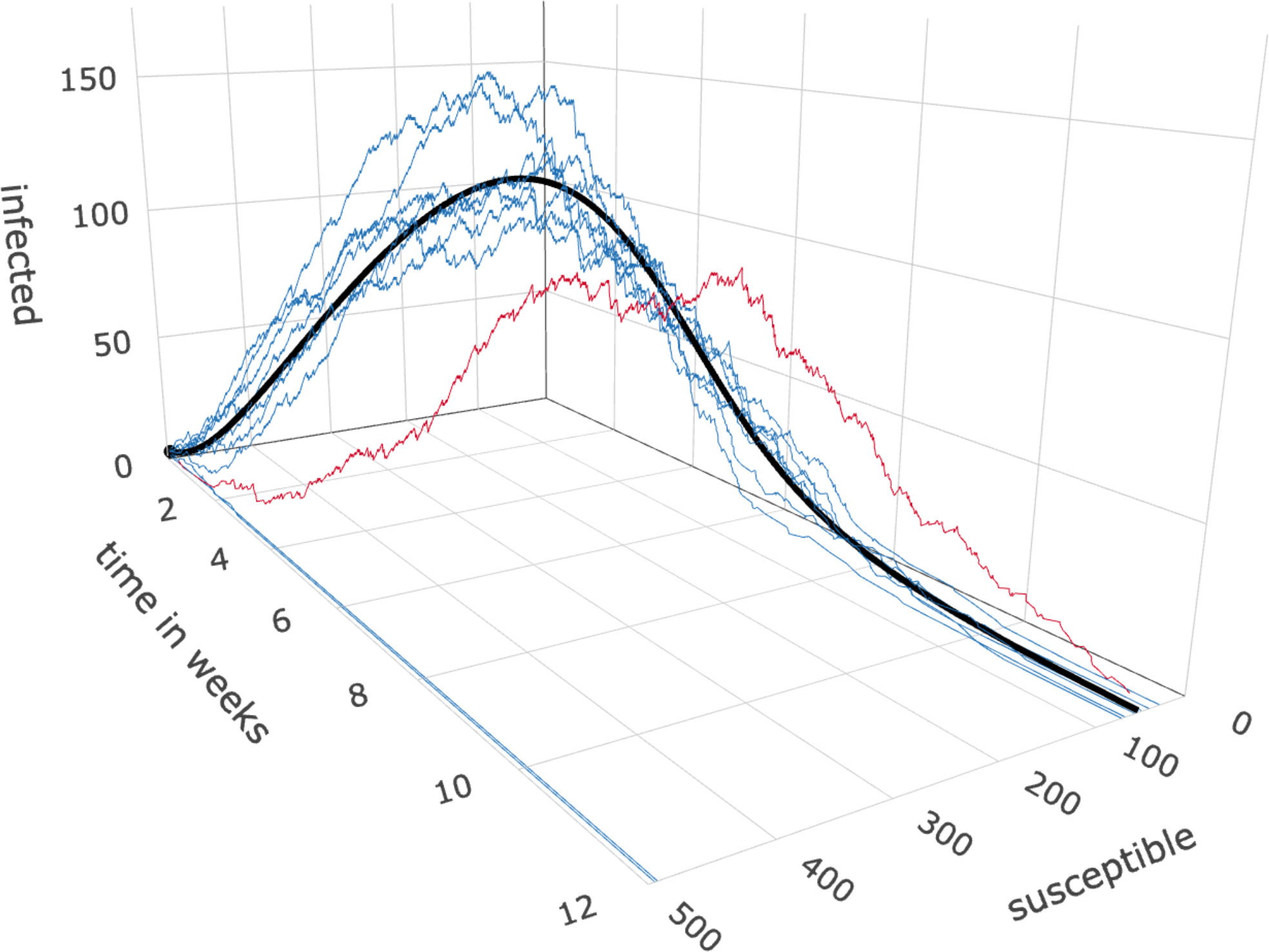}
		\caption{Solution of a deterministic SIR model (black curve) and 10 randomly sampled trajectories of the corresponding stochastic SIR model (blue and red). The trajectory highlighted in red deviates drastically from the deterministic solution.}
		\label{DeterministicPlot}
	\end{subfigure}
	\quad
	\begin{subfigure}[t]{.44\textwidth}
		\centering
		\includegraphics[width=\linewidth]{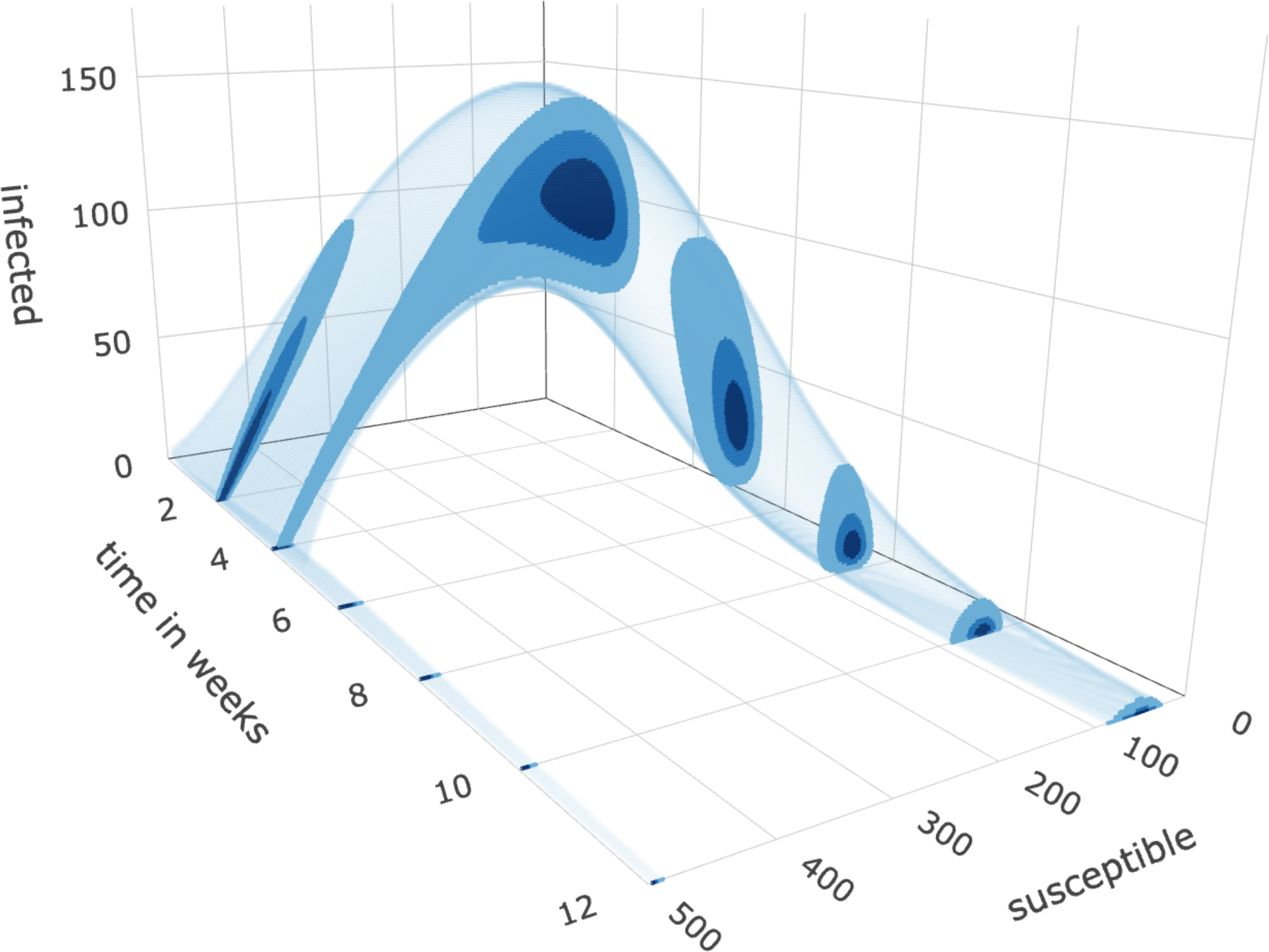}
		\caption{Analytic solution of the Kolmogorov equation for a stochastic SIR model. The time slices show distributions $\pt$ where the shades of blue show the smallest (not necessarily contiguous) areas that contain $30\%$, $60\%$ and $90\%$ of the probability mass at time $t$. }
		\label{StochasticPlot}
	\end{subfigure}
    \caption{Illustration of SIR models with $N=500$, $\alpha=1w^{-1}$, $\beta=2.5w^{-1}$, $I(0)=3$, $S(0)=497$.}
	\label{ThreeDplots}
\end{figure}
\paragraph{}
The \textbf{deterministic SIR model} \parencite{kendrick27} assumes that $S(t), I(t), R(t) \in [0,N]$ are continuous and describes their evolution over time $t \in [0,\infty)$ by the following system of nonlinear ordinary differential equations:
\begin{align} \label{eq:dsir}
\frac{\mathrm{d}S(t)}{\mathrm{d}t}   &= \overbrace{- \beta  \frac{I(t) S(t)}{N} }^{\text{infections}} \quad\overbrace{\phantom{- \alpha \frac{I(t)}{N}}}^{\text{recoveries}}, \nonumber \\
\frac{\mathrm{d}I(t)}{\mathrm{d}t}    &= + \beta  \frac{I(t) S(t)}{N}\quad- \alpha I(t), \\
\frac{\mathrm{d}R(t)}{\mathrm{d}t}   &= \phantom{+ \beta S(t) \frac{I(t)}{N}}\quad + \alpha I(t),\nonumber
\end{align}
where $\alpha, \beta \in \mathbb{R}^+$ are parameters. Note that once $S(t)$ and $I(t)$ are given, $R(t)=N-S(t)-I(t)$ is already determined and can be omitted in further analysis.

In words, an infection occurs when a susceptible person comes in sufficiently close contact with an infected person, which happens proportionally to the number of susceptible and to the density of infected people in the population and proportionally to an infection rate $\beta$. This rate $\beta$ encompasses, for example, disease characteristics, people's behavior, public policy and weather. An infected person recovers with rate $\alpha$ and can then no longer become susceptible or infected again. The basic reproduction number $\mathcal{R}_0 := \beta / \alpha$ is the number of people (in a fully susceptible population) that one infected person infects before recovering.
 
There is no analytical solution to system \eqref{eq:dsir}, but it can be solved numerically, for example by Euler's method:
\begin{align}
S(t+\Delta t) &= S(t) - \beta \frac{S(t)I(t)}{N} \Delta t, \\
I(t+\Delta t) &= I(t) + \beta \frac{S(t)I(t)}{N} \Delta t - \alpha I(t) \Delta t. \nonumber
\end{align}
The black curve in \Cref{DeterministicPlot} illustrates this solution for given parameters $\alpha=1w^{-1}$, $\beta=2.5w^{-1}$ and initial conditions $N=500$, $I(0)=3$, $S(0)=497$.

This model has several limitations. First, an epidemic is in fact a stochastic process and not a deterministic dynamical system. Second, without modeling the stochastics explicitly it is not possible to quantify the uncertainties of inferred parameters, which contributes to the uncertainties in the course of the epidemic.  

\paragraph{}
The \textbf{stochastic SIR model} \parencite{kendrick25, allen17} is a continuous-time Markov chain over all possible states of the population. A state is a pair of integers $(S,I) \in \{ 0, \dots, N \} \times \{ 0, \dots, N \}$ denoting the number of susceptible and infected people. 
States with $S+I > N$ are unreachable but still accounted for in the model.\footnote{This is necessary for the tensor representation in \cref{eq:TensorQ}.}

Let $\pt \in \mathbb{R}^{(N+1)^2}$ denote the probability distribution at time $t$ over all states $(S,I)$. That is, $\pt_{(S,I)}$ is the probability that at time $t$ there are $S$ susceptible and $I$ infected people. Its time evolution is governed by the Kolmogorov forward equation
\begin{align} \label{eq:kol}
\frac{\mathrm{d}\pt}{\mathrm{d}t} = Q \pt,
\end{align}
where the matrix $Q \in \mathbb{R}^{(N+1)^2 \times (N+1)^2}$ contains the transition rates from a state $(S,I)$ to a state $(S+\Delta S, I+\Delta I)$:
\begin{align} \label{eq:exQ}
Q_{(S+\Delta S, I+\Delta I),(S,I)} = 
\begin{cases}
\beta \frac{S I}{N} & \mbox{if } \Delta S=-1, \Delta I=+1, \\
\alpha I & \mbox{if } \Delta S= 0, \Delta I=-1, \\
-\beta \frac{S I}{N}-\alpha I & \mbox{if } \Delta S= 0, \Delta I= 0, S\neq 0, I \neq N, \\
-\alpha I & \mbox{if } \Delta S= 0, \Delta I= 0, S=0 \text{ or } I=N, \\
0 & \mbox{otherwise}.\\
\end{cases}
\end{align}
The blue and red curves in \Cref{DeterministicPlot} depict 10 randomly sampled trajectories where transitions happen according to the rates in \cref{eq:exQ}, generated by the \textcite{gillespie76} algorithm. The trajectory highlighted in red shows how stochastic fluctuations, especially at the beginning of the epidemic, can drastically alter the shape of the curve compared to its deterministic counterpart. 
\Cref{StochasticPlot} shows the analytic solution to \cref{eq:kol} and further illustrates that the stochasticity is not merely additive noise around the deterministic solution. In particular, the stochastic SIR model allows for a bifurcation where the epidemic dies out in the beginning with nonzero probability.

\begin{figure}[H]
	\centering
	\includegraphics[width=\textwidth]{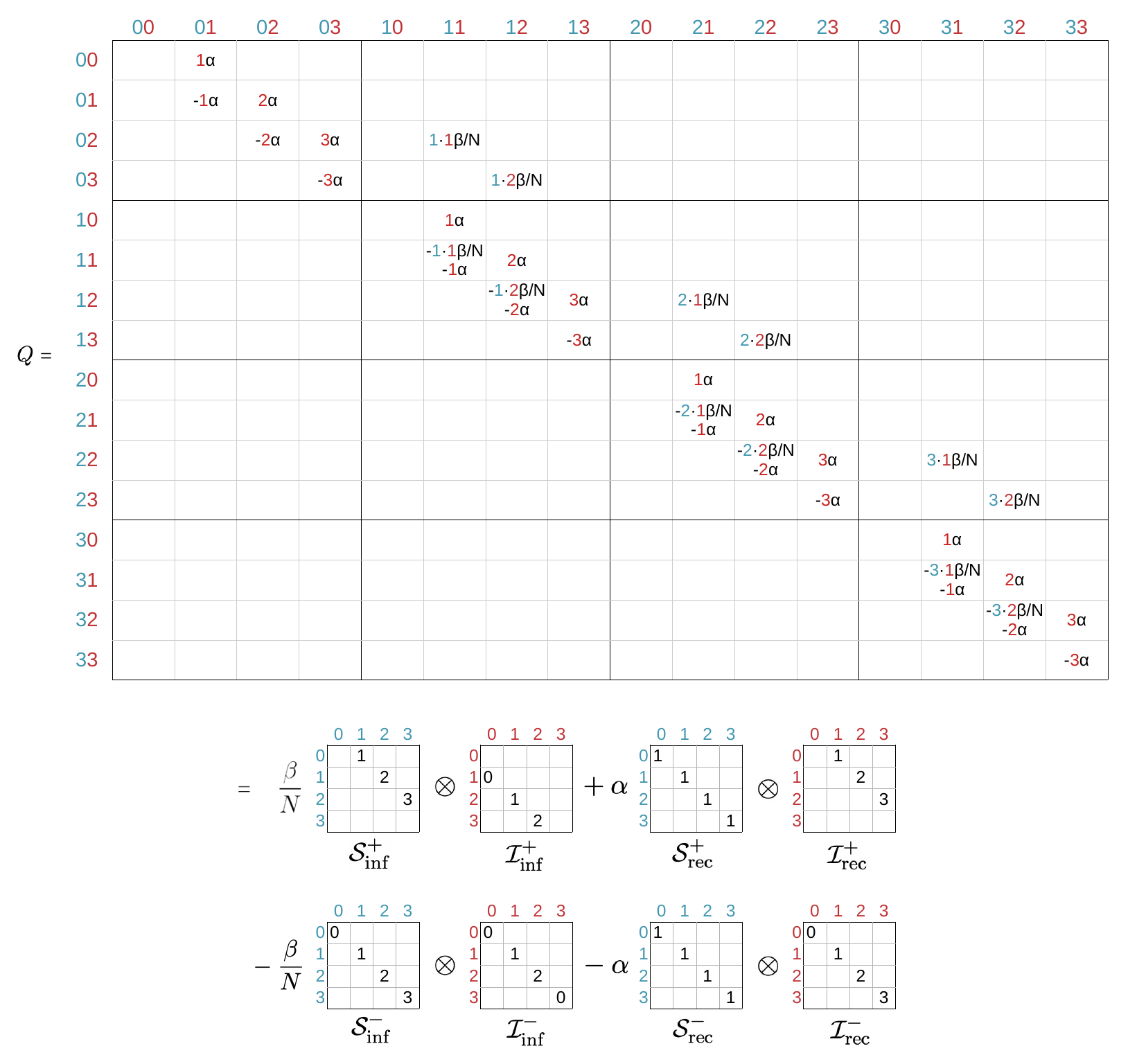}
    \captionsetup{width=.9\linewidth}
	\caption{
		\small
		Illustration of $Q$ for a population of size $N=3$ given by its entry-wise representation in \cref{eq:exQ} (top) and its tensor representation in \cref{eq:TensorQ} (bottom). Blue numbers indicate {\color[HTML]{4197B0}{susceptibles}}, red numbers indicate {\color[HTML]{C11136}{infected}} and blank entries in the matrices are zero. Transitions should be read from columns to rows.
		\\$\mathcal{S}^+_{\text{inf}}$: An infection decreases the number of susceptibles by one and happens proportionally to the current number of susceptibles.  \\$\mathcal{I}^+_{\text{inf}}$: At the same time, an infection increases the number of infected by one and happens proportionally to the current number of infected. The tensor product $\otimes$ combines both these transitions for a single infection. Moreover, an infection happens inversely proportional to the total population size $N$ and proportionally to the parameter $\beta$. \\ 
		$\mathcal{S}^+_{\text{rec}}$: A recovery does not change the number of susceptibles. \\
		$\mathcal{I}^+_{\text{rec}}$:  At the same time, a recovery decreases the number of infected by one and happens proportionally to the current number of infected. The tensor product $\otimes$ combines both these transitions for a single recovery. Moreover, a recovery happens proportionally to the parameter $\alpha$.
		\\The matrices $\mathcal{S}^-_{\text{inf}}$, $\mathcal{I}^-_{\text{inf}}$, $\mathcal{S}^-_{\text{rec}}$,
		$\mathcal{I}^-_{\text{rec}}$ generate corresponding negative entries for the diagonal of $Q$. 
	}\label{fig:CP_Illustration}
\end{figure}

The parameters $\alpha, \beta \in \mathbb{R}^+$ can be inferred from data using differentiated uniformization. This requires multiple matrix-vector products with $Q$ which is, however, too large to be stored explicitly, even for populations of only thousands of people. Hence, we propose a novel representation of $Q$ that does not require explicit storage. To this end, we introduce band matrices of size ${(N+1) \times (N+1)}$:
\begin{align}
\begin{aligned}
	\mathcal{S}^+_{\text{inf}} &= \text{superdiag}(1, \dots, N),\\
	\mathcal{S}^-_{\text{inf}} &= \text{diag}(0, \dots, N),\\
	\mathcal{S}^+_{\text{rec}} &= \text{diag}(1, 1, \dots, 1)=\Id,\\
	\mathcal{S}^-_{\text{rec}} &= \text{diag}(1, 1, \dots, 1)=\Id,
\end{aligned}
&\qquad\qquad\begin{aligned}
	\mathcal{I}^+_{\text{inf}} &= \text{subdiag}(0, \dots, N-1),\\
	\mathcal{I}^-_{\text{inf}} &= \text{diag}(0, \dots, N-1, 0),\\
	\mathcal{I}^+_{\text{rec}} &= \text{superdiag}(1, \dots, N),\\
    \mathcal{I}^-_{\text{rec}} &= \text{diag}(0, \dots, N).
\end{aligned}
\end{align} 
This yields a representation of the transition-rate matrix
\begin{align}  \label{eq:TensorQ}
Q = \frac{\beta}{N} (\mathcal{S}^+_{\text{inf}}  \otimes \mathcal{I}^+_{\text{inf}}) + \alpha (\mathcal{S}^+_{\text{rec}}  \otimes \mathcal{I}^+_{\text{rec}}) - \frac{\beta}{N} (\mathcal{S}^-_{\text{inf}}  \otimes \mathcal{I}^-_{\text{inf}}) - \alpha (\mathcal{S}^-_{\text{rec}}  \otimes \mathcal{I}^-_{\text{rec}})
\end{align}
as a sum of tensor products\footnote{For clarity, we did not factor out $\mathcal{S}^+_{\text{rec}}=\mathcal{S}^-_{\text{rec}}=\Id$, which would allow for a representation with only three terms.} (see \Cref{fig:CP_Illustration} for an illustrated explanation). Note that \cref{eq:TensorQ} is not an approximation but an exact reformulation of \cref{eq:exQ}. The benefit of this representation is that its storage complexity is $\mathcal{O}(N)$ rather than $\mathcal{O}(N^4)$ and that performing matrix-vector products has a complexity in only $\mathcal{O}(N^2)$ \parencite{buis96} rather than $\mathcal{O}(N^4)$. 

Additionally, differentiated uniformization requires the derivative $\partial Q/\partial \theta$. Here we perform inference with respect to logarithmic parameters $\theta = (\log\alpha, \log\beta)$ in order to ensure the positivity constraint on $\alpha$ and $\beta$: 
\begin{align}  \label{eq:dTensorQ}
\frac{\partial Q}{\partial\log \alpha} &= \alpha (\mathcal{S}^+_{\text{rec}}  \otimes \mathcal{I}^+_{\text{rec}}) - \alpha (\mathcal{S}^-_{\text{rec}}  \otimes \mathcal{I}^-_{\text{rec}}),\\
\frac{\partial Q}{\partial\log \beta} &= \frac{\beta}{N} (\mathcal{S}^+_{\text{inf}}  \otimes \mathcal{I}^+_{\text{inf}}) - \frac{\beta}{N} (\mathcal{S}^-_{\text{inf}}  \otimes \mathcal{I}^-_{\text{inf}}). 
\end{align}
Finally, differentiated uniformization requires a differentiable upper bound $\gamma$ on the absolute diagonal entries of $Q$. For the SIR model we choose the exact maximum 
\begin{align}  
\gamma = \underset{x}{\max}|Q_{x,x}|
&= \max\left\{ |Q_{(N-1,N-1),(N-1,N-1)}|, |Q_{(N,N),(N,N)}| \right\} \nonumber\\
&= \max\left\{ N(N-1)\frac{\beta}{N}+(N-1)\alpha, N\alpha\right\}\nonumber\\
&= \max\left\{ (N-1)\beta+(N-1)\alpha, \alpha+(N-1)\alpha\right\}\nonumber\\
&= (N-1)\alpha + \max\{(N-1)\beta, \alpha\}.
\end{align}
It is differentiable\footnote{For $\alpha = (N-1)\beta$ a differentiable upper bound for $\max\{(N-1)\beta, \alpha\}$ is $\log( e^{(N-1)\beta} + e^\alpha)$.} for $\alpha \neq (N-1)\beta$ with
\begin{align}  
\frac{\partial\gamma}{\partial \log\alpha} &= 
\begin{cases}
N\alpha  & \mbox{if } \alpha > (N-1)\beta,\\ (N-1)\alpha  & \mbox{if } \alpha < (N-1)\beta,
\end{cases}\\
\frac{\partial\gamma}{\partial \log\beta} &= 
\begin{cases}
0 & \mbox{if } \alpha > (N-1)\beta,\\ (N-1)\beta & \mbox{if } \alpha < (N-1)\beta.
\end{cases}
\end{align}
Overall, differentiated uniformization performs $\mathcal{O}(\gamma)$ matrix-vector products and thus has a total runtime complexity in $\mathcal{O}(\gamma N^2)=\mathcal{O}(N^3)$ for the SIR model. It requires storage of the result $\pt$, which has complexity $\mathcal{O}(N^2)$. 

For parameter inference we are typically only interested in the likelihood that an earlier data point $(S, I)$ is followed by a later data point $(S+\Delta S, I+\Delta I)$ after time $t$. 
Since the number of susceptibles cannot increase ($\Delta S \le 0$) and the number of recovered cannot decrease ($\Delta R = -\Delta S - \Delta I \ge 0$) along a trajectory, it is sufficient to compute $\pt$ and $\pt'$ on the restricted state space
$$\{ S+\Delta S, \dots, S \} \times \{ I -\Delta R, \dots, I-\Delta S \},$$ 
as explained in \Cref{appendixA}. Following \textcite{ho18} we use this state-space restriction to reduce the time complexity of our algorithm to $\mathcal{O}\big((I+\vert\Delta S\vert)(\Delta S^2 + \vert\Delta S\vert\Delta R)\big)$ and its storage complexity to $\mathcal{O}(\Delta S^2 + \vert\Delta S\vert\Delta R)$.

\section{COVID-19 pandemic}
\begin{figure}[H]
	\centering
	\includegraphics[width=.8\textwidth]{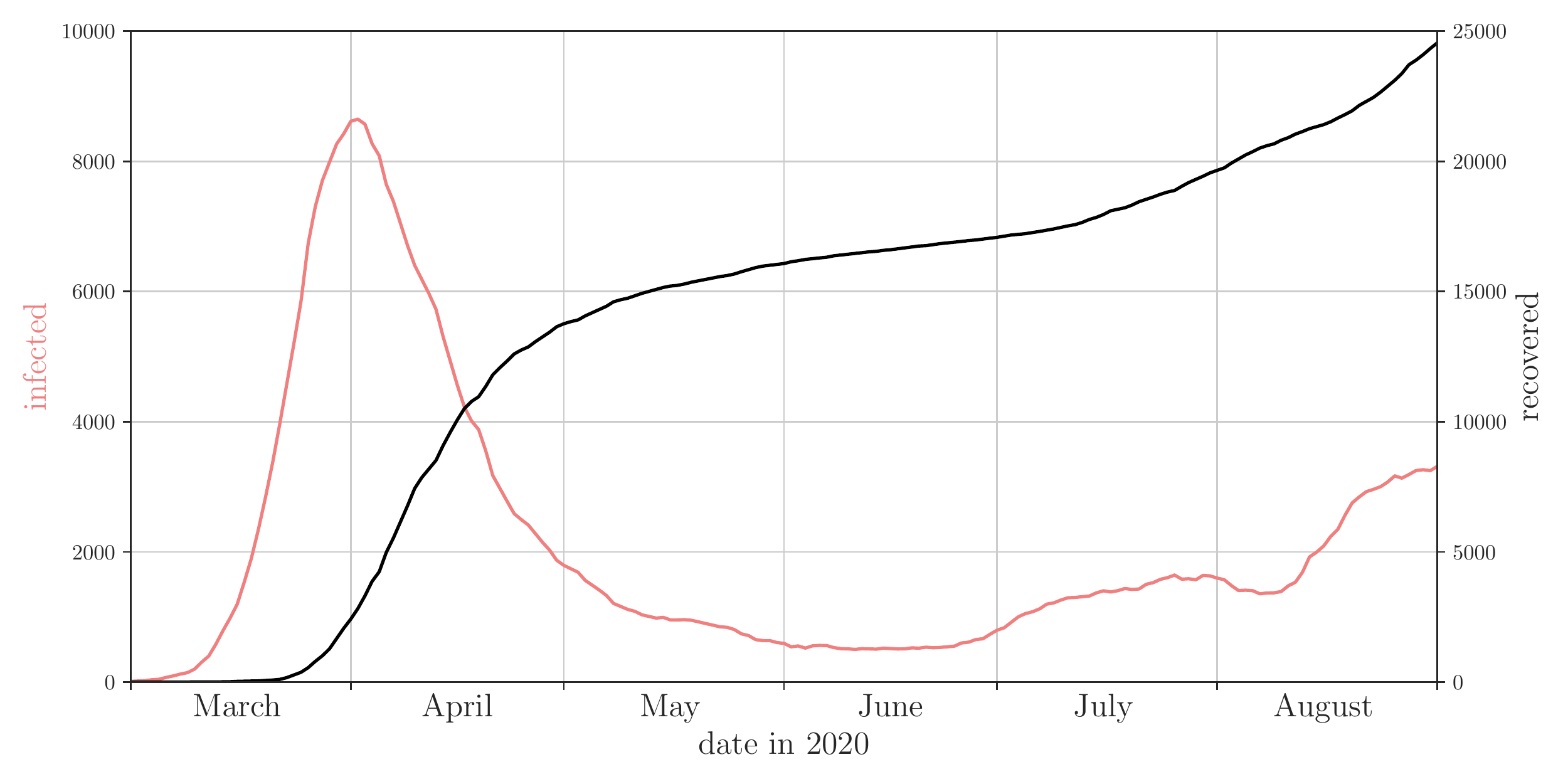}
    \captionsetup{width=.9\textwidth}
	\caption{Daily reported numbers of people infected by and recovered from SARS-CoV-2 in Austria.} 
	\label{austria_timeline}
\end{figure}
\begin{figure}[H]
\centering
\includegraphics[width=.9\textwidth]{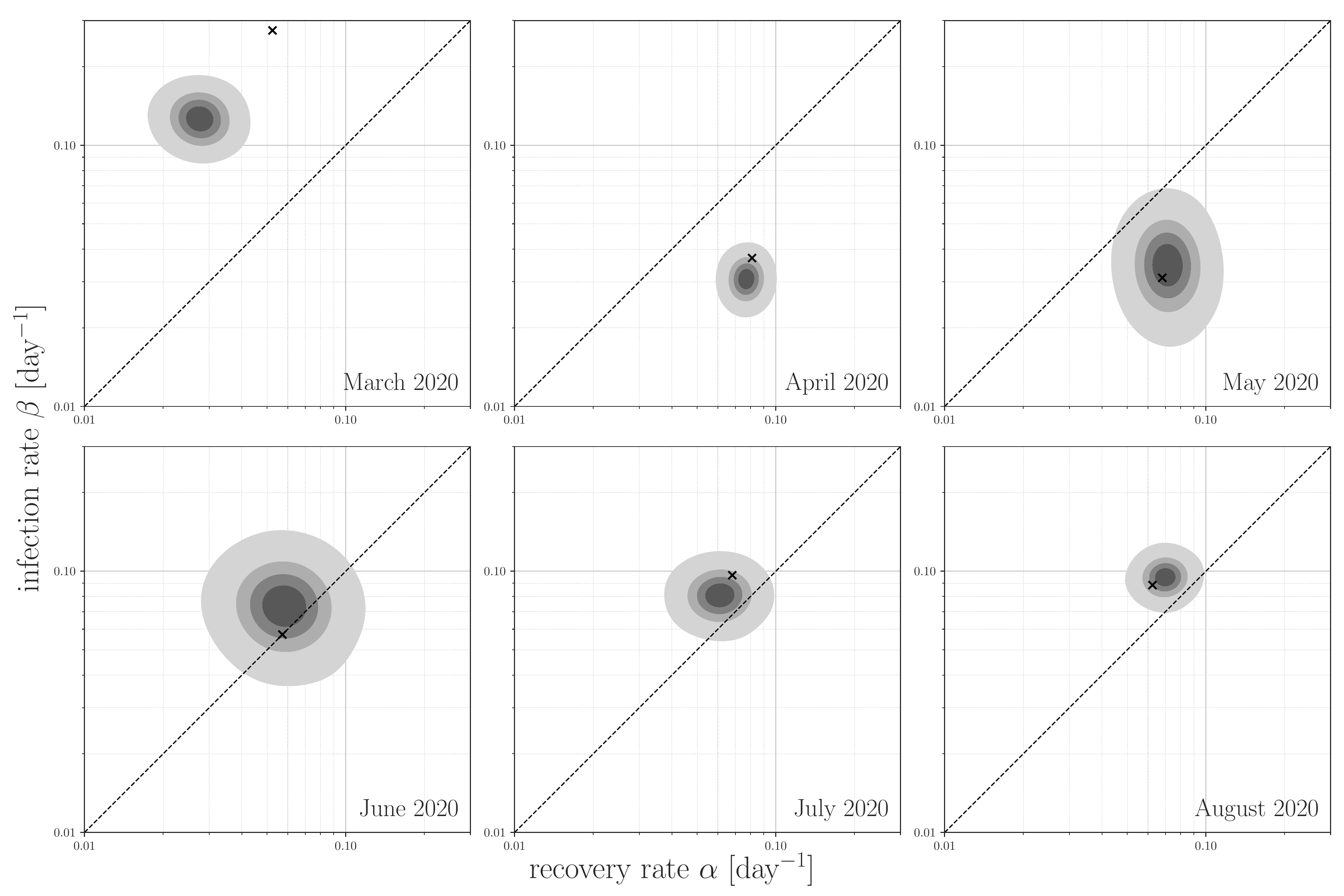}
\captionsetup{width=.9\textwidth}
\caption{Posterior probability densities over parameter pairs $(\alpha,\beta)$ for separate stochastic SIR models of the first six months of the COVID-19 pandemic in Austria. The dashed lines indicate parameters where the basic reproduction number $\mathcal{R}_0=\beta/\alpha=1$. The crosses mark the least-squares estimators of the corresponding deterministic SIR models.}
\label{HMCplots}
\end{figure}

Here we model the first wave of the COVID-19 pandemic in Austria as a stochastic SIR model. We employ differentiated uniformization to estimate the parameters $\alpha$ and $\beta$ and quantify their uncertainty. 
We use daily numbers on $S$, $I$ and $R$ between 2020-03-01 and 2020-09-01 from public health data provided by the Austrian \textcite{bmsgpk} (\Cref{austria_timeline}). $I$ and $R$ are given directly, and we set $S=N-I-R$ assuming that the initial population size $N=8{,}932{,}664$ stays constant. People who have died from COVID-19 are counted under ``recovered'' in a technical sense as they are no longer infectious.
We do not correct for undiscovered cases and biases in testing and reporting. We also assume that parameters are piecewise constant for each month.

We do a full Bayesian analysis for parameter pairs $(\log\alpha,\log\beta)$ with a uniform prior. Following \textcite{ho18} we sample from the joint posterior using a Hamiltonian Monte Carlo (HMC) scheme \parencite{duane87,neal11}. Unlike a standard Metropolis-Hastings scheme, HMC makes use of the gradient of the likelihood, which we compute using differentiated uniformization. This makes sampling more efficient with less samples needed to cover the posterior distribution \parencite{gelman04}. We estimated the joint posterior of $(\log\alpha,\log\beta)$ for every month between March 2020 and August 2020 separately. For each month we performed 10 parallel Monte Carlo chains with length 100, where we discarded the first 10 points each, resulting in 900 points per month. These calculations were done on the QPACE 4 cluster \parencite{georg21} and took about 30 minutes for each of May and June, 2 hours for July, 16 hours for each of April and August and 4 days for March.

\Cref{HMCplots} shows the results of this analysis. The estimated posterior is plotted for $(\alpha,\beta)$ on logarithmic scales. The gray shaded areas were generated using Gaussian-kernel density estimation applied to the posterior samples. The crosses mark the least-squares estimators of the corresponding deterministic SIR models. The dashed lines represent parameter constellations where $\alpha=\beta$ and thus $\mathcal{R}_0=1$. Here the epidemic switches between growing and decreasing numbers of infected. From April-August 2020 the posterior of the recovery rate $\alpha$ varies
around a value of $0.07$ per day, corresponding to the realistic mean time to recovery of about 2 weeks \parencite{faes20}. In contrast, the posterior of $\alpha$ in March 2020 appears to be off, with a mean of about $0.03$ per day corresponding to a mean time to recovery of one month.
Inspecting the original numbers, we observed that the numbers of recovered are unexpectedly low (less than 100 people until 2020-03-23) possibly due to lagging declaration of recoveries because of cautious hospital policies in the beginning of the pandemic. 

Overall, we observe large uncertainties associated with the parameters in several but not all months. These might hint at epidemic courses that are not perfectly in line with an SIR model or with the assumption of piecewise constant parameters.
Such deviations from the model assumptions are much less readily apparent in deterministic approaches.

\section{Discussion}
We provide a novel method for computing the transient distribution and its derivative for continuous-time Markov Chains on huge discrete state spaces. This makes parameter inference tractable for a large family of statistical models, including the stochastic SIR model of epidemic spread.

Our key observation is that the transition-rate matrix of an SIR model can be written as a sum of tensor products, which allows us to cheaply compute matrix-vector products without storing the matrix itself. This operation alone is sufficient to compute the transient distribution by the uniformization method \parencite{grassmann77}, a numerically stable power-series expansion of the matrix exponential. We propose the \emph{differentiated uniformization method}, an analogous power series for computing the derivative of the transient distribution with respect to parameters of a CTMC.

For the SIR model our algorithm scales cubically in the size of the population, which is one order slower than the state-of-the-art method for multivariate birth processes \parencite{ho18}. On the other hand, our general-purpose algorithm also applies to birth-death processes such as predator-prey dynamics \parencite{owen14}, which have been considered intractable so far \parencite{ho17}. We illustrate this in \Cref{appendixB}.

Beyond epidemiology we are interested in tumor progression modeling using mutual hazard networks \parencite{schill19}. 
Similar to an epidemiological model that scales exponentially in the number of compartments, a tumor progression model is a CTMC that scales exponentially in the number of possible mutations. 
While both differentiated uniformization and the algorithm of \textcite{ho18} have the potential to advance this field, large scale inference remains an open problem for tumor progression models with up to hundreds of mutations.
The tensor representation of the transition-rate matrix could serve as a starting point for representing the transient distribution itself in a low-rank tensor format.
These formats reduce the exponential cost (e.g., in the number of mutations or compartments) to linear cost provided certain low-rank structures exist \parencite{hackbusch12}.
For large-scale CTMCs, low-rank tensor formats were already successfully used, e.g., for the computation of transient \parencite{johnson10} and stationary distributions \parencite{benson17,buchholz16,kressner14} and also for a variant of the uniformization method \parencite{klever20}. 
Therefore, the combination of low-rank tensor formats and differentiated uniformization could be a promising new avenue for large-scale inference problems in computational oncology and epidemiology.

From this perspective our work can also be seen as an attempt to connect these two communities.

\section{Acknowledgements}
This work was funded by the German Research Foundation (DFG) grants TRR-305, SFB/TRR-55 and GR-3179/6-1.

\printbibliography

\appendix
\section{State-space restriction}\label{appendixA}

In order to compute the likelihood that an earlier data point  $(S, I)$ is followed by a later data point $(S+\Delta S, I+\Delta I)$ after time $t$, it is sufficient to compute $\pt$ on a small subset of the entire state space. Since the number of susceptibles cannot increase ($\Delta S \leq 0$) and the number of recovered cannot decrease ($\Delta R = -\Delta S - \Delta I \geq 0$), all possible trajectories from $(S, I)$ to $(S+\Delta S, I+\Delta I)$ must necessarily stay within the restricted state space
$$\{ S_{\text{min}}, \dots, S_{\text{max}} \} \times \{ I_{\text{min}}, \dots, I_{\text{max}} \},$$
where
\begin{align}
S_{\text{min}} = S + \Delta S, &\qquad S_{\text{max}} = S, \nonumber\\
I_{\text{min}} = I - \Delta R, &\qquad I_{\text{max}} = I - \Delta S.
\end{align}

All probability mass that leaves this space must be accounted for, but we do not need to keep track of its destination. To this end, we introduce the modified band matrices
\begin{align} 
\underbrace{\begin{aligned}
	\tilde{\mathcal{S}}^+_{\text{inf}} &= \text{superdiag}(S_{\text{min}}+1, \dots, S_{\text{max}})\\
	\tilde{\mathcal{S}}^-_{\text{inf}} &= \text{diag}(S_{\text{min}}, \dots, S_{\text{max}})\\
	\tilde{\mathcal{S}}^+_{\text{rec}} &= \text{diag}(1, 1, \dots, 1)=\Id\\
	\tilde{\mathcal{S}}^-_{\text{rec}} &= \text{diag}(1, 1, \dots, 1)=\Id\\
\end{aligned}}_{\begin{matrix}(\vert\Delta S\vert + 1)\times(\vert\Delta S\vert + 1)\end{matrix}}
&\qquad\qquad\underbrace{\begin{aligned}
	\tilde{\mathcal{I}}^+_{\text{inf}} &= \text{subdiag}(I_{\text{min}}, \dots, I_{\text{max}} - 1)\\
	\tilde{\mathcal{I}}^-_{\text{inf}} &= \text{diag}(I_{\text{min}}, \dots, I_{\text{max}})\\
	\tilde{\mathcal{I}}^+_{\text{rec}} &= \text{superdiag}(I_{\text{min}} + 1, \dots, I_{\text{max}})\\
	\tilde{\mathcal{I}}^-_{\text{rec}} &= \text{diag}(I_{\text{min}}, \dots, I_{\text{max}}).
&\end{aligned}}_{\begin{matrix}(\vert\Delta S\vert + \Delta R + 1)\times(\vert\Delta S\vert + \Delta R + 1)\end{matrix}}
\end{align} 
and define a smaller transition-rate matrix on the restricted state space as 
\begin{align}  
\tilde{Q} &= \frac{\beta}{N} (\tilde{\mathcal{S}}^+_{\text{inf}}  \otimes \tilde{\mathcal{I}}^+_{\text{inf}}) + \alpha (\tilde{\mathcal{S}}^+_{\text{rec}}  \otimes \tilde{\mathcal{I}}^+_{\text{rec}}) - \frac{\beta}{N} (\tilde{\mathcal{S}}^-_{\text{inf}}  \otimes \tilde{\mathcal{I}}^-_{\text{inf}}) - \alpha (\tilde{\mathcal{S}}^-_{\text{rec}}  \otimes \tilde{\mathcal{I}}^-_{\text{rec}})
\end{align}
with derivatives
\begin{align}
\frac{\partial \tilde{Q}}{\partial\log \alpha} &= \alpha (\tilde{\mathcal{S}}^+_{\text{rec}}  \otimes \tilde{\mathcal{I}}^+_{\text{rec}}) - \alpha (\tilde{\mathcal{S}}^-_{\text{rec}}  \otimes \tilde{\mathcal{I}}^-_{\text{rec}}),\\
\frac{\partial \tilde{Q}}{\partial\log \beta} &= \frac{\beta}{N} (\tilde{\mathcal{S}}^+_{\text{inf}}  \otimes \tilde{\mathcal{I}}^+_{\text{inf}}) - \frac{\beta}{N} (\tilde{\mathcal{S}}^-_{\text{inf}}  \otimes \tilde{\mathcal{I}}^-_{\text{inf}}). 
\end{align}
Note that the columns of $\tilde{Q}$ sum to less than zero and that $\pt$ therefore sums to less than 1 on the restricted state space. Computing matrix-vector products using these operators has a time complexity in $\mathcal{O}(\vert\Delta S\vert^2 + \vert\Delta S\vert\Delta R)$.

The largest absolute diagonal entry of $\tilde{Q}$ is 
\begin{align}
\gamma = \underset{x}{\max}|\tilde{Q}_{x,x}|  = 
 \frac{\beta}{N} S_\text{max} I_\text{max} + \alpha I_\text{max} 
\end{align}
with derivatives
\begin{align}
\frac{\partial\gamma}{\partial \log\alpha} &= \alpha I_\text{max}, \\
\frac{\partial\gamma}{\partial \log\beta} &=  \frac{\beta}{N} S_\text{max} I_\text{max}.
\end{align}

We perform $m$ iterations of \cref{algo2} such that the entire probability mass (including that which left the restricted state space) according to \cref{eq:MassDefect} reaches the required tolerance. Hence, the overall time complexity of the algorithm is
\begin{align}
\mathcal{O}\big(\gamma(\vert\Delta S\vert^2 + \vert\Delta S\vert\Delta R)\big)=\mathcal{O}\big(I_{\text{max}}(\vert\Delta S\vert^2 + \vert\Delta S\vert\Delta R)\big)=\mathcal{O}\big((I+\vert\Delta S\vert)(\vert\Delta S\vert^2 + \vert\Delta S\vert\Delta R)\big).
\end{align}
Storing the result $\pt$ has complexity $\mathcal{O}(\vert\Delta S\vert^2 + \vert\Delta S\vert\Delta R)$.

\section{Predator-prey dynamics} \label{appendixB}
\begin{figure}[H]
	\centering
	\begin{subfigure}[t]{.44\textwidth}
		\centering
		\includegraphics[width=\linewidth]{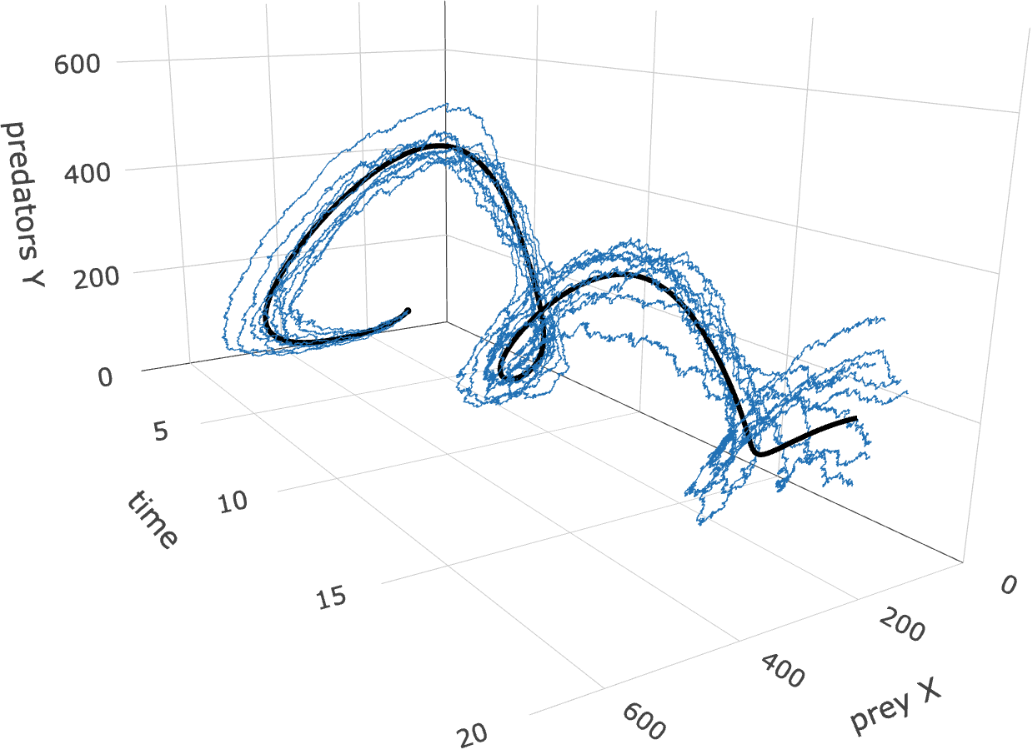}
		\caption{Solution of a deterministic predator-prey model (black curve) and 10 randomly sampled trajectories (blue) of the corresponding stochastic model.}
		\label{DeterministicPlotLV}
	\end{subfigure}
	\quad
	\begin{subfigure}[t]{.44\textwidth}
		\centering
		\includegraphics[width=\linewidth]{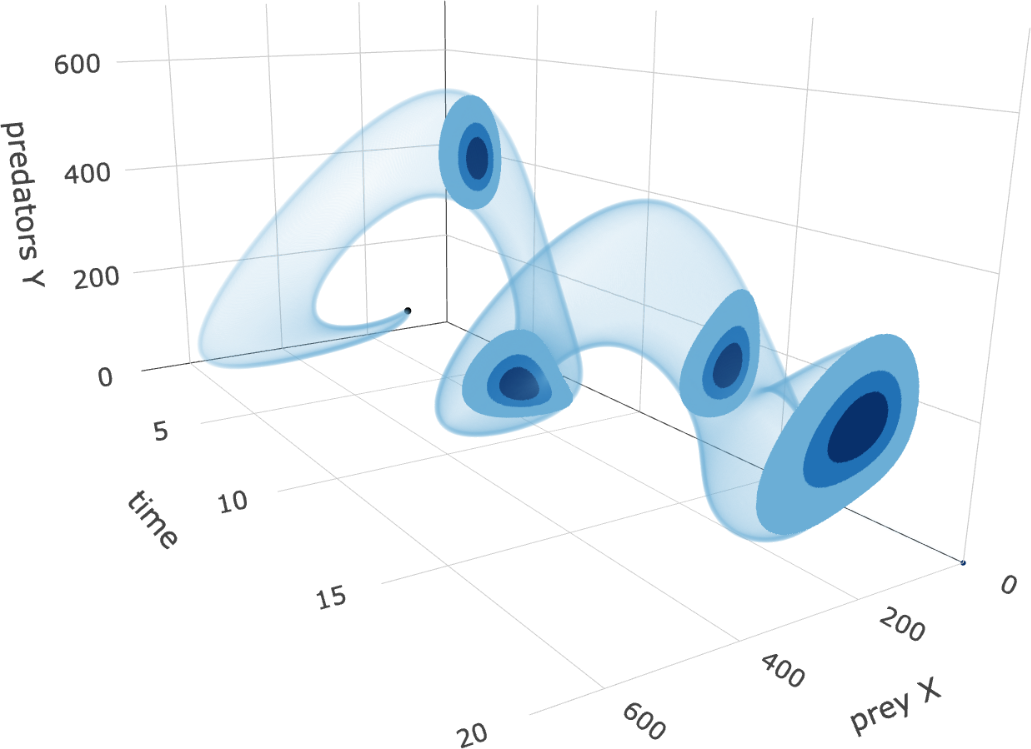}
		\caption{Analytic solution of the Kolmogorov equation for a stochastic predator-prey model. The time slices show distributions $\pt$ where the shades of blue show the smallest areas that contain $30\%$, $60\%$ and $90\%$ of the probability mass at time $t$. }
		\label{StochasticLV}
	\end{subfigure}
    \captionsetup{width=.9\textwidth}
    \caption{Illustration of predator-prey models with $ \alpha=1$, $\beta=0.004$, $\delta=0.8$, $X(0)=100$, $Y(0)=40$, $X_\text{max}=Y_\text{max}=1200$.}
	\label{LV}
\end{figure}

We consider the following deterministic predator-prey equations, based on \parencite{lotka25, volterra26}, 
\begin{align} 
\frac{\mathrm{d}X(t)}{\mathrm{d}t}   &=
{\phantom{.}-\beta X(t) Y(t)} \enspace
\overbrace{\phantom{.}+\alpha X(t) - \alpha \frac{X(t)^2}{X_\text{max}}}^{\text{prey birth}} \phantom{\quad - \delta Y(t)} \\
\frac{\mathrm{d}Y(t)}{\mathrm{d}t}   &= 
\underbrace{\phantom{.}+\beta X(t) Y(t)}_{\substack{\text{prey consumption} \\ \text{\& predator birth}}}\enspace 
\phantom{\alpha X(t) - \alpha \frac{X(t)^2}{X_\text{max}}}
\enspace \underbrace{\phantom{.}- \delta Y(t)}_{\text{predator death}}
\end{align}
which describe how the population size $X(t)  \in \mathbb{R}^+$ of a prey species and the population size $Y(t) \in \mathbb{R}^+$ of a predator species change continuously over time as the species interact (black curve in \Cref{DeterministicPlotLV}). The parameter $\alpha$ is the birth rate of prey, $\delta$ is the death rate of predators and $\beta$ is the contact rate between predators and prey. Upon contact a prey is consumed and, we assume for simplicity, exactly one predator is born. 
$X_\text{max}$ is a finite carry capacity representing the available plant resources for the prey species, which would result in logistic growth in the absence of predators. 
This is neither necessary nor commonly assumed in the literature on the deterministic model, since the prey population is always limited by a nonzero number of predators in $\mathbb{R}^+$.

Here we are interested in a corresponding stochastic model \parencite{owen14} (blue curves in \Cref{DeterministicPlotLV}) in which the number of predators is an integer and may drop to zero, which would lead to exponential growth of the prey population without finite carry capacity.
We define a stochastic predator-prey model as a CTMC over the state space 
$$ \{0, \dots, X_\text{max}\} \times \{0, \dots, Y_\text{max} \}$$
with transition-rate matrix
\begin{align} \label{eq:exQLV}
Q_{(X+\Delta X, Y+\Delta Y),(X,Y)} = 
\begin{cases}
\beta XY & \mbox{if } \Delta X=-1, \Delta Y=+1, \\
\delta Y & \mbox{if } \Delta X= 0, \Delta Y=-1, \\
\alpha X-\alpha X^2/X_\text{max} & \mbox{if } \Delta X= +1, \Delta Y=0, \\
-\beta XY -\delta Y - \alpha X+\alpha X^2/X_\text{max}  & \mbox{if } \Delta X= 0, \Delta Y= 0,\\
0 & \mbox{otherwise}\\
\end{cases}
\end{align}
whose columns sum to less than zero. This is because the upper limit $Y_\text{max}$ of the predator population is a computational cutoff and not enforced by the model. Transitions that leave the state space result in missing probability mass, which must be mitigated by choosing a sufficiently high $Y_\text{max}$.

We introduce the band matrices
\begin{align}
\underbrace{\begin{aligned}
\mathcal{X}^+_{\text{cons}} &= \text{superdiag}(1, \dots, X_\text{max}),\\
\mathcal{X}^-_{\text{cons}} &= \text{diag}(0, \dots, X_\text{max}),\\
\mathcal{X}^+_{\text{birth}} &= \text{subdiag}(0, \dots, X_\text{max}-1),\\
\mathcal{X}^-_{\text{birth}} &= \text{diag}(0, \dots, X_\text{max}-1, 0),\\
\mathcal{X}^+_{\text{cap}} &= \text{subdiag}(0^2, 1^2, 2^2, \dots, (X_\text{max}-1)^2),\\
\mathcal{X}^-_{\text{cap}} &= \text{diag}(0^2, 1^2, 2^2, \dots, (X_\text{max}-1)^2, 0),\\
\mathcal{X}^+_{\text{death}} &= \text{diag}(1, 1, \dots, 1)=\Id,\\
\mathcal{X}^-_{\text{death}} &= \text{diag}(1, 1, \dots, 1)=\Id,\\
\end{aligned}}_{\begin{matrix}(X_\text{max}+1)\times(X_\text{max}+1)\end{matrix}}
&\qquad\qquad\underbrace{\begin{aligned}
\mathcal{Y}^+_{\text{cons}} &= \text{subdiag}(0, \dots, Y_\text{max}-1),\\
\mathcal{Y}^-_{\text{cons}} &= \text{diag}(0, \dots, Y_\text{max}),\\
\mathcal{Y}^+_{\text{birth}} &= \text{diag}(1, 1, \dots, 1)=\Id,\\
\mathcal{Y}^-_{\text{birth}} &= \text{diag}(1, 1, \dots, 1)=\Id,\\
\mathcal{Y}^+_{\text{cap}} &= \text{diag}(1, 1, \dots, 1)=\Id,\\
\mathcal{Y}^-_{\text{cap}} &= \text{diag}(1, 1, \dots, 1)=\Id,\\
\mathcal{Y}^+_{\text{death}} &= \text{superdiag}(1, \dots, Y_\text{max}),\\
\mathcal{Y}^-_{\text{death}} &= \text{diag}(1, \dots, Y_\text{max}),\\
\end{aligned}}_{\begin{matrix}(Y_\text{max}+1)\times(Y_\text{max}+1)\end{matrix}}
\end{align} 
in order to represent the transition-rate matrix
\begin{align}  
Q = &+\beta (\mathcal{X}^+_{\text{cons}}  \otimes \mathcal{Y}^+_{\text{cons}}) + \alpha(\mathcal{X}^+_{\text{birth}}  \otimes \mathcal{Y}^+_{\text{birth}}) -\frac{\alpha}{X_\text{max}}(\mathcal{X}^+_{\text{cap}}  \otimes \mathcal{Y}^+_{\text{cap}}) +\delta(\mathcal{X}^+_{\text{death}}  \otimes \mathcal{Y}^+_{\text{death}})\notag\\
&-\beta(\mathcal{X}^-_{\text{cons}}  \otimes \mathcal{Y}^-_{\text{cons}}) - \alpha (\mathcal{X}^-_{\text{birth}}  \otimes \mathcal{Y}^-_{\text{birth}}) +\frac{\alpha}{X_\text{max}}(\mathcal{X}^-_{\text{cap}}  \otimes \mathcal{Y}^-_{\text{cap}}) - \delta(\mathcal{X}^-_{\text{death}}  \otimes \mathcal{Y}^-_{\text{death}})
\end{align}
as a sum of tensor products.
This allows us to efficiently compute solutions $\pt$ of the Kolmogorov equation for the stochastic predator-prey model (see \Cref{StochasticLV}). 
\end{document}